\documentclass[letterpaper]{article} 
\usepackage[]{aaai23}  
\usepackage{times}  
\usepackage{helvet}  
\usepackage{courier}  
\usepackage[hyphens]{url}  
\usepackage{graphicx} 
\usepackage{natbib}  
\usepackage{caption} 
\usepackage{algorithm}
\usepackage{algorithmic}
\usepackage{tikz}
\usepackage{array}
\usepackage{amsmath}
\usepackage{amssymb}
\usepackage{bbm}
\newcolumntype{P}[1]{>{\centering\arraybackslash}p{#1}}

\usepackage{newfloat}
\usepackage{listings}
\DeclareCaptionStyle{ruled}{labelfont=normalfont,labelsep=colon,strut=off} 
\floatstyle{ruled}
\newfloat{listing}{tb}{lst}{}
\floatname{listing}{Listing}
\title{Launchpad: Learning to Schedule Using Offline and Online RL Methods}

\author{
    Vanamala Venkataswamy, \textsuperscript{\rm 1}
    Jake Grigsby, \textsuperscript{\rm 2}
    Andrew Grimshaw, \textsuperscript{\rm 3}
    Yanjun Qi \textsuperscript{\rm 1}
}
\affiliations{
    \textsuperscript{\rm 1}University of Virginia, Charlottesville, Virginia 22904 USA\\

    \textsuperscript{\rm 2}University of Texas at Austin, Austin, Texas 78712 USA\\

    \textsuperscript{\rm 3}Lancium Compute, Charlottesville, Virginia 22902 USA\\
}

\begin{document}

\maketitle

\begin{abstract}

Deep reinforcement learning algorithms have succeeded in several challenging domains. Classic Online RL job schedulers can learn efficient scheduling strategies but often takes thousands of timesteps to explore the environment and adapt from a randomly initialized DNN policy. Existing RL schedulers overlook the importance of learning from historical data and improving upon custom heuristic policies. Offline reinforcement learning presents the prospect of policy optimization from pre-recorded datasets without online environment interaction. Following the recent success of data-driven learning, we explore two RL methods: 1) Behaviour Cloning and 2) Offline RL, which aim to learn policies from logged data without interacting with the environment. These methods address the challenges concerning the cost of data collection and safety, particularly pertinent to real-world applications of RL. Although the data-driven RL methods generate good results, we show that the performance is highly dependent on the quality of the historical datasets. Finally, we demonstrate that by effectively incorporating prior expert demonstrations to pre-train the agent, we short-circuit the random exploration phase to learn a reasonable policy with online training. We utilize Offline RL as a \textbf{launchpad} to learn effective scheduling policies from prior experience collected using Oracle or heuristic policies. Such a framework is effective for pre-training from historical datasets and well suited to continuous improvement with online data collection.

\end{abstract}

\section{Introduction and Motivation}

Reinforcement Learning (RL) has solved sequential decision tasks of impressive difficulty by maximizing reward functions through trial and error. Recent examples using deep learning range from robotic locomotion~\cite{robotics_2015}, sophisticated video games~\cite{games_2018}, congestion control~\cite{congestion_control}, and job scheduling~\cite{Vana_Rare-JSSPP2022}. 

The process of reinforcement learning involves iteratively interacting with the environment and collecting experience, typically with the most recently learned policy, and then using the experience to improve the policy ~\cite{sutton-barto-2018}. In many settings, this online interaction is impractical because data collection is expensive (e.g., robotics, healthcare) or dangerous (e.g., autonomous driving). Moreover, even in domains where online interaction is possible (e.g., job scheduling), practitioners prefer utilizing previously collected data, especially if the domain is complex and requires large datasets for effective generalization. 


\begin{figure*}[t]
\centering
\includegraphics[width=\textwidth]{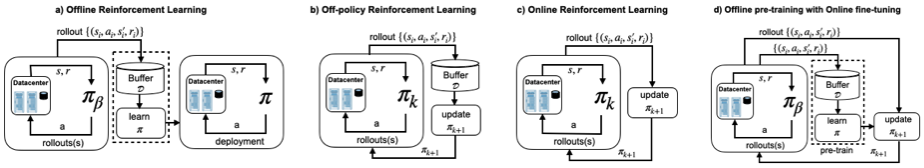}
\caption{Illustration of offline reinforcement learning (a), classic off-policy reinforcement learning (b), classic online reinforcement learning (c), and Offline training with Online fine-tuning (d).}
\label{fig:RL_BC_Offline}
\end{figure*}


Furthermore, learning a task from scratch can require a prohibitively time-consuming amount of exploration of the state-action space to find a good policy, especially in sparse reward environments. Additionally, existing RL schedulers overlook the importance of learning and improving upon existing heuristic policies. Learning without prior knowledge is an approach rarely taken in the natural world. Knowledge of how to approach a new task can be transferred from previously learned tasks or extracted from the performance of an expert. The RL agents can leverage the behavior of custom heuristic policies explicitly designed for unique environments to learn and improve overall performance. The heuristic policies generate expert demonstrations, and the RL agents learn from these demonstrations to improve upon the heuristic policies.

For many of these domains, including job scheduling, large amounts of historical data are readily available. Effective data-driven methods for deep reinforcement learning (DRL) should be able to utilize this data to pre-train offline while improving with online fine-tuning. This has led to a resurgence of interest in data-driven RL methods, namely 1) Behaviour Cloning (BC) and 2) Offline RL (historically known as batch RL)~\cite{batch-rl}, which aim to learn policies from logged data without further interaction with the real system. 

Behavior cloning is an approach for imitation learning~\cite{pomerleau}, where the policy is trained with supervised learning to imitate the actions of a provided dataset directly. This process is highly dependent on the performance of the data-collecting process. In many cases, these come from an existing rule-based, heuristic, or myopic policy that we are trying to replace with an RL approach. The effective use of such datasets would make real-world RL more practical and enable better generalization by incorporating diverse prior experiences. Due to the efficient use of collected data and the stability of the learning process, this research area has attracted much attention recently. 

Depending on the quality of the prior demonstrations, useful knowledge can be extracted about the task being solved, the dynamics of the environment, or both. Pure BC methods incorporate prior data with the aim of directly mimicking demonstrations. This is desirable, assuming demonstrations are known to be optimal. However, it enforces strict requirements on offline data quality, which can cause undesirable bias when the demonstration data is not optimal. Often, the collected data can be mixed with sub-optimal transitions. 

Offline reinforcement learning algorithms also aim to leverage large existing datasets of previously collected data to produce effective policies that generalize across a wide range of scenarios without needing costly active data collection. Empirical studies comparing offline RL to imitation learning have come to mixed conclusions. Some studies show that offline RL methods outperform imitation learning significantly, specifically in environments that require “stitching” parts of suboptimal trajectories~\cite{d4rl}. In contrast, many recent articles have argued that BC performs better than offline RL on both expert and suboptimal demonstration data over a variety of tasks~\cite{implicitBC}~\cite{hahn}~\cite{offlineRL-what-matters}. This makes it confusing for practitioners to understand whether to use offline RL or merely run BC on collected demos. 

While offline learning methods provide a mechanism for utilizing prior data, such methods are generally ineffective for fine-tuning online data as they are often too conservative. In effect, these methods require us to question: Do we assume the prior dataset is optimal? Do we use strictly offline data or only online data? We need algorithms that learn successfully in either of these cases to make it feasible to learn policies for real-world settings. Therefore, we study a simple actor-critic algorithm that bridges pre-training from prior data and improvement with online data collection. This RL technique is effective for pre-training from off-policy datasets but is also well suited to continuously improve with online data collection.

Our \textbf{contribution} in this paper is the empirical characterization of BC, Offline RL, and Online RL pre-trained with offline datasets for job scheduling in the green datacenter environment. The contribution of this study is not just another RL scheduler but a systematic study of what makes sense - standard offline RL, pure Online RL methods, or offline pre-training with online fine-tuning. We evaluate these RL techniques in a simulated green datacenter environment where the RL scheduler agent is responsible for effectively scheduling jobs on available resources.

\section{Background} \label{background}

Figure \ref{fig:RL_BC_Offline} illustrates offline reinforcement learning (a), classic off-policy reinforcement learning (b), classic online reinforcement learning (c), and Offline training with Online fine-tuning (d).

\textbf{Reinforcement learning (RL).} RL is a framework aimed at dealing with tasks of sequential nature. Typically, the problem is defined as a Markov decision process (MDP) $(S, A, R, p, \gamma)$, with state space S, action space A, reward function R (scalar), transition dynamics p, and discount factor $\gamma$. The behavior of an RL agent is determined by a policy $\pi$. The RL agent's objective is to maximize the long-term expected discounted return $E_{\pi}[\sum_{t=0}^{\infty} \gamma^t r_{t+1}]$, i.e., the expected cumulative sum of rewards when following the policy in the MDP. This objective is evaluated by a value function, which measures the expected discounted return after taking action a in state s: $Q^{\pi}(s, a) = E_{\pi} [\sum_{t=0}^{\infty} \gamma^t r_{t+1} \vert s_0 = s, a_0 = a]$. 

\textbf{Behavior Cloning (BC).} Another approach for training policies is imitating an expert or behavior policy. Behavior cloning (BC) is an approach for imitation learning~\cite{pomerleau}, where the policy is trained with supervised learning to imitate the actions of a provided dataset directly. Unlike RL, the success of BC is highly dependent on the quality of the dataset. BC is likely to fail when the prior dataset does not contain enough transitions generated by a policy performing well on the task, or the signal-to-noise ratio is too large.

\textbf{Off-policy.} Off-policy learning consists of a behavior policy that generates the data and a target policy that learns from this data~\cite{sutton-barto-2018}. The behavior policy continuously collects data for the agent in the environment. For example, data is collected using previous policies up to time $k$ during training $\pi_0, \pi_1, \ldots, \pi_k$ and stored in a replay buffer. This data is used to train the policy $\pi_{k+1}$. An example of off-policy RL is actor-critic variants~\cite{espeholt18a}. 

\textbf{Offline RL.} Offline RL (a.k.a batch RL) breaks the presumption that the agent can interact with the environment. Instead, we provide the agent with a fixed dataset collected by some unknown data-generating process (experts or heuristic policies). This setting may become challenging since the agent loses the opportunity to explore the environment according to its current views and must infer good behavior from only the provided transitions. More generally, offline RL is a counterfactual inference problem: given data generated from a given set of decisions infer the consequence of an independent set of decisions.

\section{Green Datacenter - Overview} \label{design}

The green datacenter is a datacenter co-located at or near renewable energy sources. One or more renewable sources can power the datacenter with the provision to store (batteries) excess energy from renewables. Additionally, the datacenter is connected to the electric grid to support critical infrastructure when energy from renewables and batteries cannot sustain the load. Our datacenter design is akin to the model presented in ~\cite{Vana_Rare-JSSPP2022}. 

\begin{figure}[h]
\centering
\includegraphics[width=\linewidth]{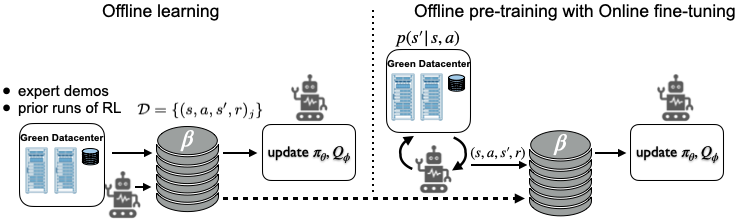}
\caption{RL scheduling agent - Offline Learning (left) and Offline pre-training with Online fine-tuning (right).} 
\label{fig:rl_gdc}
\end{figure}

The goal is to design a green datacenter environment controlled by RL and heuristic-based scheduling policies. In order to train the RL scheduler, we convert the datacenter scheduling problem into an MDP with a state space $\mathbf{S}$ representing the current status of the datacenter resources, an action space $\mathbf{A}$ of new jobs, and a reward function $\mathbf{R}$ to be maximized. The operation of the datacenter - including accepting new jobs and placing jobs on available resources - becomes the MDP transition function. Figure \ref{fig:rl_gdc} provides an overview of an Offline agent learning from prior datasets and Online scheduler agent - pre-trained with Offline data - fine-tuning the policy by actively interacting with the green datacenter environment. 

\subsection{State Space, Actions, and Rewards} 
The state space, $\mathcal{S}$, includes information about resources, jobs, and resource availability (based on power generation predictions). 

\subsubsection{Resources} 
Our green datacenter simulator models a pool of servers (CPUs and GPUs), enabling the scheduler to make granular per-resource scheduling decisions. The resource pool size expands and contracts based on the power available at the datacenter at any given time. Power availability decides when and how many resources are turned on or off. As the power availability fluctuates, the corresponding resource availability is reflected in the state information supplied to the scheduler agent. 

\subsubsection{Jobs} 
In our system, jobs can be in one of three locations: 1) wait\_pool, 2) ready\_pool, or 3) scheduled on the resources. The wait\_pool is where jobs first arrive. The jobs from wait\_pool are moved to the ready\_pool, where they can then be scheduled on the resources. Jobs have meta-data, including the job's id, value, QoS, and resource requirements. The jobs are processed over fixed T timesteps. The time-horizon shifts after processing jobs during that timestep, with the job metadata vectors updated and the resource image advancing by one row. As the time-horizon shifts, the power supply from renewables dictates the availability of resources. The scheduler agent continuously observes the state of jobs, resources, and resource/power availability to make scheduling decisions. 

\textit{QoS of a job}: Users are willing to pay different amounts for different jobs based on their importance. The QoS value is specified as a percentage of the time the user wants his job to run. The qos\_violation\_time, $\ (expected\_finish\_time \div Qos\ Value$), specifies the upper bound by which the job must finish executing. If a job remains in the system past qos\_violation\_time, it incurs negative rewards. 

\subsubsection{Actions} 
The action space for a datacenter with ready\_pool size $n$ is a set of $n+2$ discrete options $\mathcal A = \{j_0, j_1, \ldots ,j_n, suspend, no\_op\}$. The actions $\{a=j_i, \forall i \leq n\}$ schedule the $i$th ready\_pool job $j_i$ on available resources. The action $a=suspend$ is used to suspend an incomplete job and replace it with a higher-value job. Finally, the action $a=no\_op$ means that the scheduler agent does not want to schedule (e.g., insufficient resources) or suspend any jobs in that timestep.

\subsubsection{Rewards} 
Rewards are a combination of the positive reward (job completion) or associated cost (QoS violation) for the action in a given state. Rewards are sparse in that the agent receives a reward only after the scheduled jobs complete execution. Even a small improvement in total job value can generate millions of dollars in savings for the service providers. Our RL scheduler's objective is to maximize the total job value from finished jobs, $Total\ Job\ Value = \sum_{i=1}^{\vert J_{finished}\vert} j_i.value$.

\section{Scheduling: Offline and Online methods} 

A datacenter with a ready\_pool of size $n$ is converted into an RL environment with $n + 2$ discrete actions that simulates job scheduling and returns a new state representation and reward. States are tuples containing both the resource image and array of job metadata, while the reward function can be adjusted to reflect the goals of our scheduling system. This paper focuses on optimizing the total value (revenue) from completed jobs; the reward at timestep $t$ is the total value of all completed jobs. 

The agent learns to select jobs from the ready\_pool that maximize total job value with the help of three neural networks. The \textit{encoder} combines the state information in the resource allocation image and job metadata array and produces a compact vector representation. The resource allocation image is processed by convolutional layers common in computer vision applications, while the job array is passed through standard feed-forward layers. The representations of both input types are concatenated and merged by additional feed-forward layers. The resulting state vector serves as the input to the RL agent's \textit{actor} and \textit{critic} networks. The core RL algorithm used to train the encoder, actor, and critic varies depending on whether we use online or offline data. The Offline RL algorithm is described in Algorithm \ref{algo:offline_training}, and the Online RL algorithm is described in Algorithm \ref{algo:online_training}. 

\begin{algorithm} [tbh]
	\caption{Offline training process} 
    \label{algo:offline_training}
    
	\textbf{Input}: {Advantage Samples $k$, Replay Buffer with pre-provided transitions $\mathcal{D} \leftarrow \{(s_i, a_i, r_i, s_i'), \dots\}$, function $f$} \\
    \textbf{Init}: {Actor net $\pi_{\theta}$, Critic net $Q_{\phi}$ } 
    \begin{algorithmic}[1]
    \FOR {training step $t \in \{0,\dots,T\}$}
    {
        \STATE Sample Batch of $B$ transitions $\{(s_i, a_i, r_i, s_i')\}_{i=0}^{i=B}$ from $\mathcal{D}$ \\
        \STATE {actor loss (see \cite{wang2020critic})} \\
        \STATE $\mathcal{L}_{a} \leftarrow \displaystyle \frac{1}{B} \sum_{i=0}^{i=B} f(Q_{\phi}, s_i, a_i)\text{log} \pi_{\theta}(a_i \mid s_i) $ \\
        \STATE {critic loss} \\
        \STATE $\mathcal{L}_{c} \leftarrow \displaystyle \frac{1}{B} \sum_{i=0}^{i=B} \left(\left(Q_{\phi}(s_i, a_i) - \mathop{\mathbb{E}}\left[(r_i + \gamma Q_{\phi}(s_i', a')\right]\right)^2\right)$ \\
        \STATE {update nets by gradient descent}
        \STATE $\phi \leftarrow \displaystyle \phi - \alpha \nabla{\phi}\mathcal{L}_{c}$  \\
        \STATE $\theta \leftarrow \displaystyle \theta - \alpha \nabla_{\theta}\mathcal{L}_{a}$ 
    }
    \ENDFOR
    \STATE \textbf{Output}: Trained Scheduling Policy $\pi_{\theta}(s)$ 
    \end{algorithmic}
\end{algorithm}

Both methods train a Q-function critic with the standard mean-squared Bellman error update \cite{sutton-barto-2018}, where our critic network(s), $Q_{\phi}$, learn to predict bootstrapped estimates of future value based on immediate rewards and a ``target" network. We implement an ensemble method that trains multiple critics to reduce over-estimation error, as in SAC \cite{haarnoja2018soft} and TD3 \cite{fujimoto2018addressing}. We use PopArt \cite{hessel2019multi} normalization to standardize the magnitude of our critic outputs; this simplifies tuning hyperparameters across datacenter environments with different resources and job values. 

\begin{algorithm} [tbh]
    \caption{Online training process with Offline data} 
    \label{algo:online_training}
    
	\textbf{Input}: Datacenter Simulator Env with Dynamics $\mathbf{T} : \mathcal{S} \times \mathcal{A} \rightarrow \mathcal{S}$, Reward Function $\mathbf{R} : \mathcal{S} \times \mathcal{A} \times \mathcal{S} \rightarrow \mathbb{R}$, Replay Buffer $\mathcal{D} \leftarrow \{(s_i, a_i, r_i, s_i'), \dots\} $ \\
    \textbf{Init}: Actor net $\pi_{\theta}$, Critic net $Q_{\phi}$
    \begin{algorithmic}[1]
    \FOR {training step $t \in \{0,\dots,T\}$}
        \STATE{sample an action from the policy} \\
        \STATE $a_t \sim \pi_{\theta}(s_t)$ \\
        \STATE {advance datacenter sim and receive next state and reward} \\
        \STATE $s_t' \leftarrow \mathbf{T}(s_t, a_t),$ \\
        \STATE $r_t \leftarrow \mathbf{R}(s_t, a_t, s_t')$ \\
        \STATE {add transition to the replay buffer}
        \STATE $\mathcal{D} \leftarrow \mathcal{D} \cup \{(s_t, a_t, r_t, s_t')\}$ \\
        \STATE Randomly Sample Batch of $B$ transitions $\{(s_i, a_i, r_i, s_i')\}_{i=0}^{i=B} \sim \mathcal{D}$ \\
        \STATE {critic loss} \\
        \STATE $\mathcal{L}_{c} \leftarrow \displaystyle \frac{1}{B} \sum_{i=0}^{i=B} \left(\left(Q_{\phi}(s_i, a_i) - \mathop{\mathbb{E}}\left[(r_i + \gamma Q_{\phi}(s_i', a')\right]\right)^2\right)$ \\
        \STATE{online actor loss (see \cite{haarnoja2018soft})} \\
        \STATE $\mathcal{L}_{a} \leftarrow \displaystyle \frac{1}{B} \sum_{i=0}^{i=B} \left(\mathop{\mathbb{E}}_{a' \sim \pi_{\theta}(s_i)}\left[-Q_{\phi}(s_i, a')\right]\right)$ \\
        \STATE {update nets by gradient descent} \\
        \STATE $\phi \leftarrow \displaystyle \phi - \alpha \nabla{\phi}\mathcal{L}_{c}$ \\
        \STATE $\theta \leftarrow \displaystyle \theta - \alpha \nabla_{\theta}\mathcal{L}_{a}$ \\
    \ENDFOR
    \STATE \textbf{Output}: Trained Scheduling Policy $\pi_{\theta}(s)$
    \end{algorithmic}
\end{algorithm}

The primary difference between online and offline variants is the gradient update of the actor. The online algorithm trains the actor to maximize the value predictions of the critic, as in the discrete-action version of SAC \cite{christodoulou2019soft}. Like SAC, we use a max-entropy RL formulation that encourages robust policies and prevents our agents' actions from collapsing to a local optima during training. As an additional exploration measure, we use an epsilon-greedy strategy \cite{mnih2015human} to add random noise to action selection during the early stages of learning. Directly maximizing the outputs of the critic exploits overestimation error in unfamiliar state-action pairs \cite{kumar2019stabilizing}. This ``uninformed optimism" can be a useful exploration strategy for online learning but becomes a significant issue during offline learning when we are unable to evaluate actions' true value. The offline actor update needs to constrain the policy to the distribution of actions covered by the existing training data. This is achieved by performing supervised learning on the state-action mapping contained in the training data, filtered by a function $f$ (Algo \ref{algo:offline_training}, line 4). 

Intuitively, the actor network is trying to copy the decisions in the dataset, but the filter $f$ prioritizes some actions over others. When $f$ outputs the same value for all samples, we recover standard Behavioral Cloning as a special case. A popular alternative is to up-weight actions that lead to a higher return than the current policy and down-weight those that lead to a lower return. This concept is formalized by the advantage function $A^{\pi}(s, a) = Q^{\pi}(s, a) - \mathbb{E}_{a' \sim \pi(s)}[Q^{\pi}(s, a')]$. In practice, the advantage is estimated using the critic network and samples from the actor: $A^{\pi_{\theta}}(s, a) \approx Q_{\phi}(s, a) - \frac{1}{k}\sum_{j=1}^{j=k}Q_{\phi}(s, a_j \sim \pi_{\theta}(s))$. The filter function $f$ can then down-weight samples with low advantage, such as $f(Q_{\phi}, s, a) = \text{exp}(A^{\pi_{\theta}}(s, a))$ \cite{nair2020accelerating}. Our work follows \cite{wang2020critic} and uses a more intuitive binary filter that simply ignores data that the critic network does not think will improve the policy $f(Q_{\phi}, s, a) = \mathbbm{1}\{A^{\pi_{\theta}}(s, a) > 0\}$.

\section{Experimental Results} 

\subsection{Baseline Policies}
The baseline heuristic scheduling policies are Shortest-Job-First (SJF), First-Come-First-Serve (FCFS), QoS (Quality of Service), and Highest Value First (HVF). The SJF, FCFS, and HVF have established heuristic policies and intuitive definitions. The QoS scheduling policy schedules job with the highest QoS value. Empirically, we found that the QoS heuristic policy performs best compared to other policies, with SJF as second best for small and medium problem sizes (10 to 100 resources). We ran five training runs for each learning method with randomly initialized scheduler agents. To evaluate, we averaged ten rollouts (for each of 5 training runs) with 100K steps and random seed. The 95\% confidence interval for each point in the following graphs is within $\pm 0.015-0.023$.

\subsection{Workload}
We used a synthetic workload where each job consists of meta-data, including job-id, resource requirement, and job duration. Synthetic workload provides more nuanced control over simulation parameters allowing us to study the scheduler's behavior under a wide range of conditions~\cite{keynote-jsspp}. Jobs arrive in an online fashion, meaning that the scheduler does not know the job information \textit{a priori}. The job arrival rate is controlled by $\lambda$; the higher the jobs arrival rate, the more jobs there are to process. The job arrival rate of 120\% is used for all the experiments. The performance metric used is Total Job Value.

\subsection{Offline Learning: BC and Offline RL}
\textbf{Behavior Cloning (BC)}: We start by demonstrating the performance of the first data-driven method called Behavior Cloning. The performance of BC methods is highly dependent on the quality of the training dataset. BC is likely to fail to learn good policy when the dataset does not contain enough transitions generated by a well-performing policy or the fraction of poor data is too large. For this experiment, we collected transitions from heuristic policies, namely SJF, QoS, FCFS, and HVF policies. We then load a combination of these transitions into the replay buffer BC\_combo scheduler agent. Each policy contributes 25\% of the total transitions loaded into the replay buffer. We repeat the training by loading only QoS and SJF transitions into the replay buffer to train the BC\_qos\_sjf agent. In this case, each heuristic policy contributes 50\% of the total transitions loaded into the replay buffer. Finally, we loaded only QoS transitions into the replay buffer, i.e., 100\% of the replay buffer consists of transitions from QoS policy to train the BC\_qos agent.

\begin{figure}[tbh]
\centering
\includegraphics[scale=0.50]{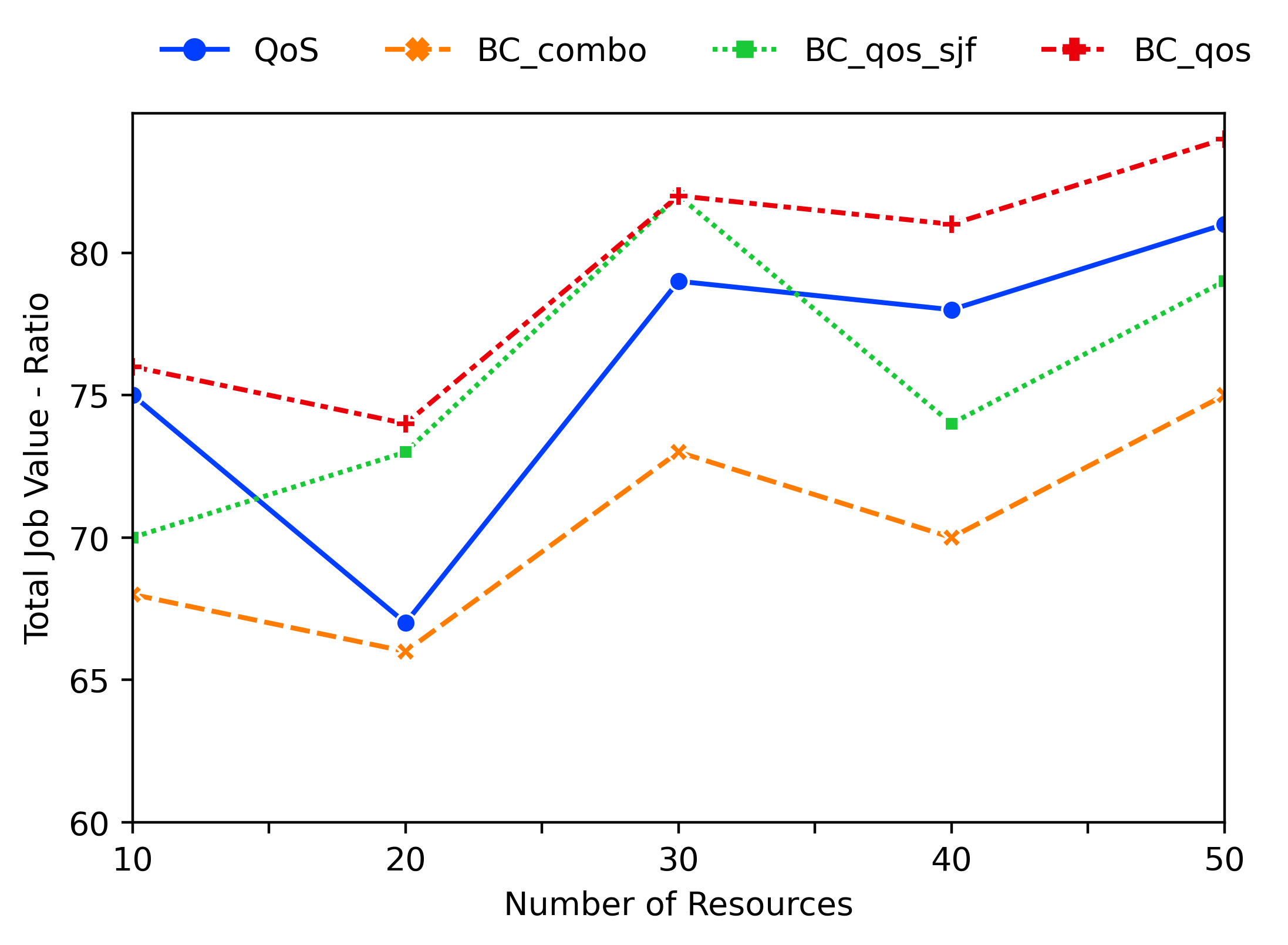}
\caption{Behaviour Cloning with heuristic policies.} 
\label{fig:BC comparision}
\end{figure}

\textit{Analysis}: Figure \ref{fig:BC comparision} shows the performance of BC agents when trained with different ratios of transitions from different heuristics. When the replay buffer consists of a mix of expert and poor quality transitions (e.g., QoS, SJF, FCFS, HVF), the BC method (BC\_combo) learns a policy that performs worse than the QoS policy alone. Even with a 50\% noise-to-signal ratio (i.e., 50\% QoS and SJF), BC\_qos\_sjf performs poorly in some instances. When the reply buffer is populated with a higher ratio of good transitions (i.e., 100\% QoS, best heuristic policy), then BC\_qos effectively learns to mimic that policy. The BC method performs significantly better when trained with high-quality demonstrations. 

We note that BC learning methods are incredibly beneficial in two scenarios. First, suppose we have prior datasets from an Oracle or a good heuristic policy but need to learn a general policy that works in a similar but slightly different setting. In that case, BC can learn the general policy from transitions generated by the Oracle for the new setting. Second, suppose we have transitions from a legacy system and cannot or do not have access to the legacy system that generated the transitions. In that case, BC learning can be applied to learn the underlying policy from the transitions generated by the legacy system.

\textbf{Offline RL}: Next, we demonstrate the performance of the second data-driven method called Offline RL. Unlike BC, the performance of Offline RL is resilient to training datasets with mixed (both well-performing and poor) heuristic policies. Suppose prior datasets do not contain enough transitions generated by a well-performing policy, or the fraction of poor data is too large. In that case, Offline RL methods can leverage the benefits of stitching parts of suboptimal trajectories. Similar to the previous experiment, we load a combination of heuristic transitions (SJF, QoS, FCFS, and HVF policies) into the replay buffer for training the Offline RL agents.

\begin{figure}[tbh]
\centering
\includegraphics[scale=0.50]{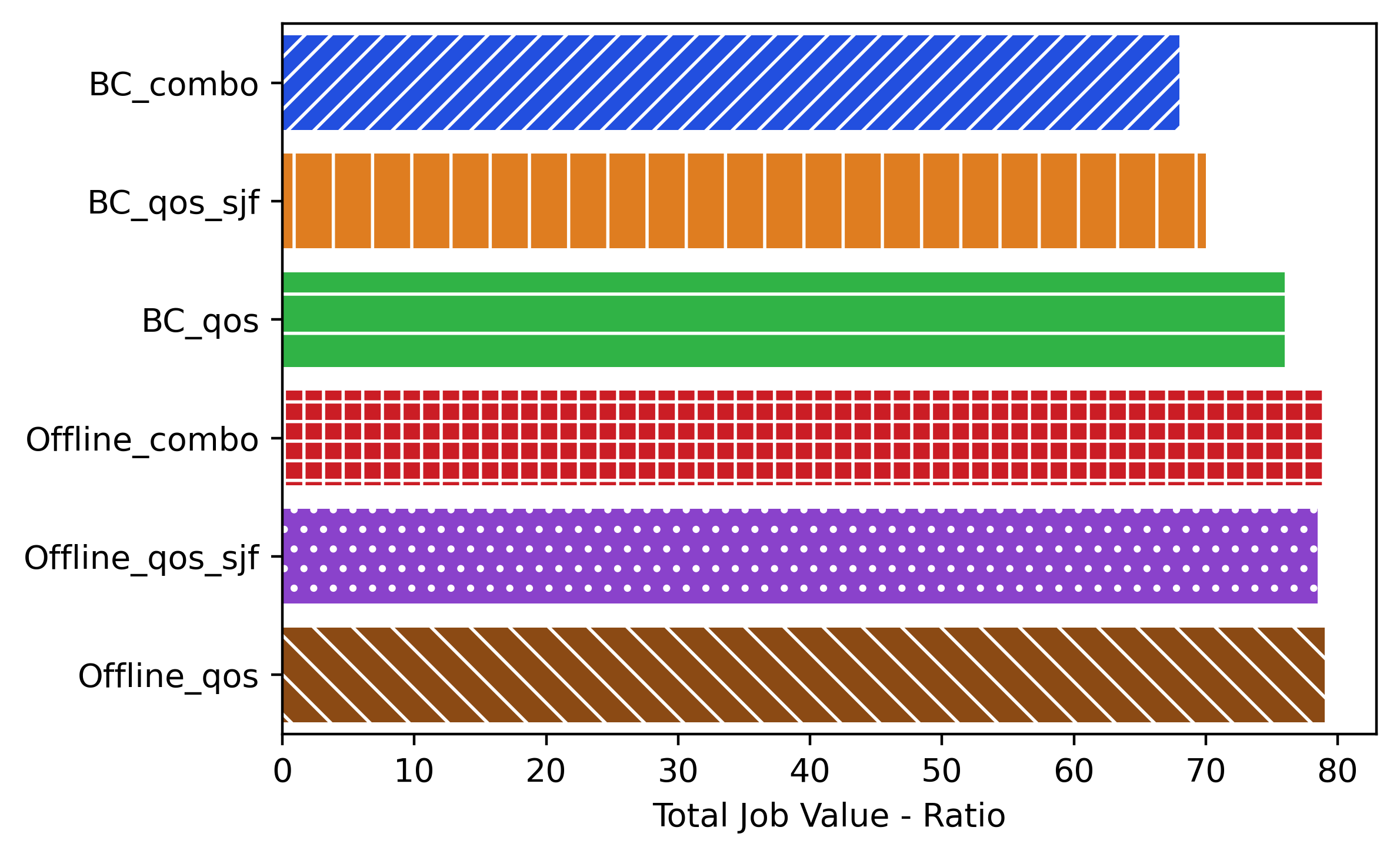}
\caption{Comparing Offline Learning and Behaviour Cloning with combination of heuristic policies.} 
\label{fig:Offline_RL_10Res}
\end{figure}

\begin{figure}[tbh]
\centering
\includegraphics[scale=0.50]{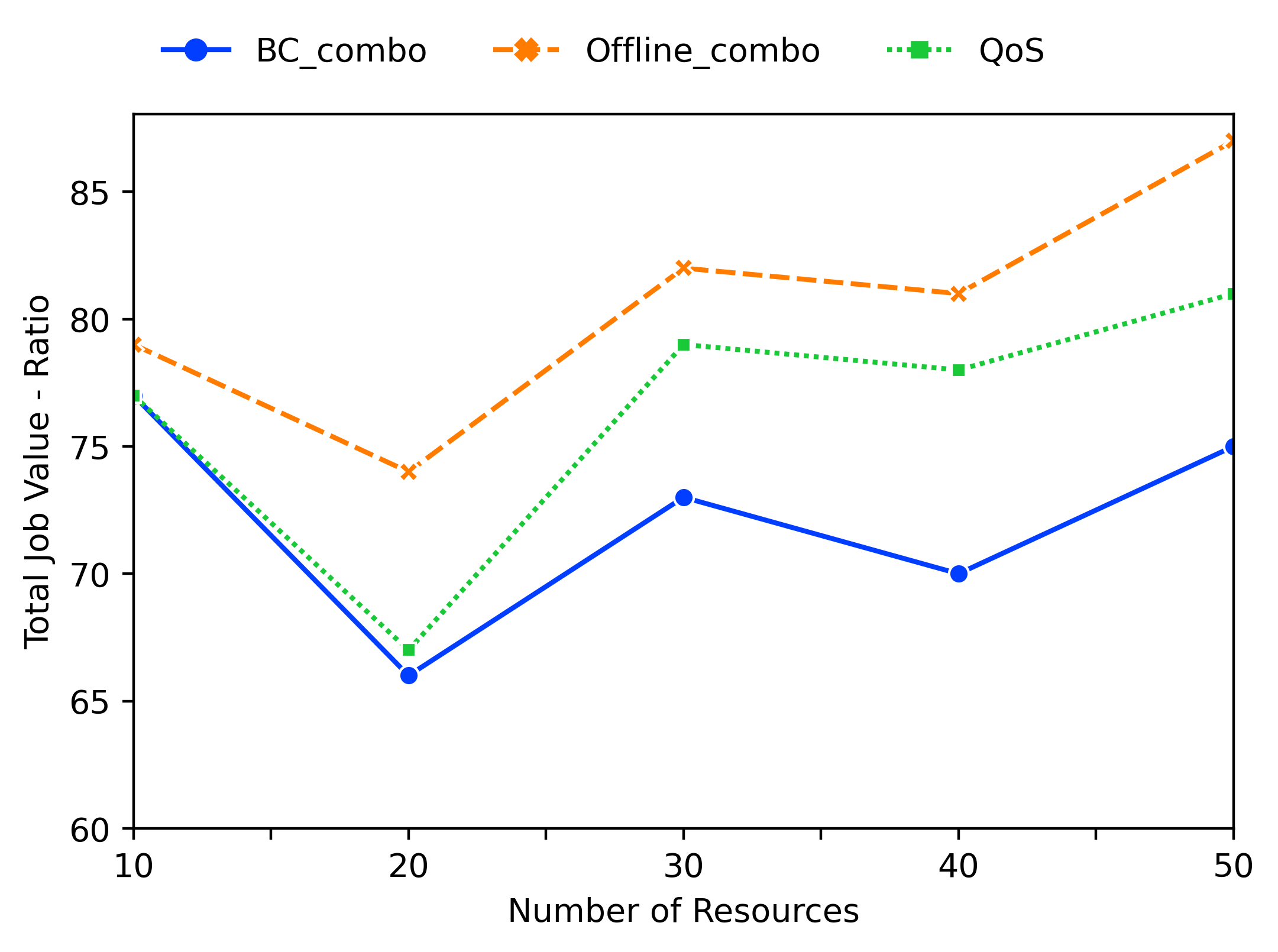}
\caption{Comparing Offline Learning and Behaviour Cloning with combination of heuristic policies.} 
\label{fig:Offline_RL_scale}
\end{figure}

\begin{figure}[tbh]
\centering
\includegraphics[scale=0.50]{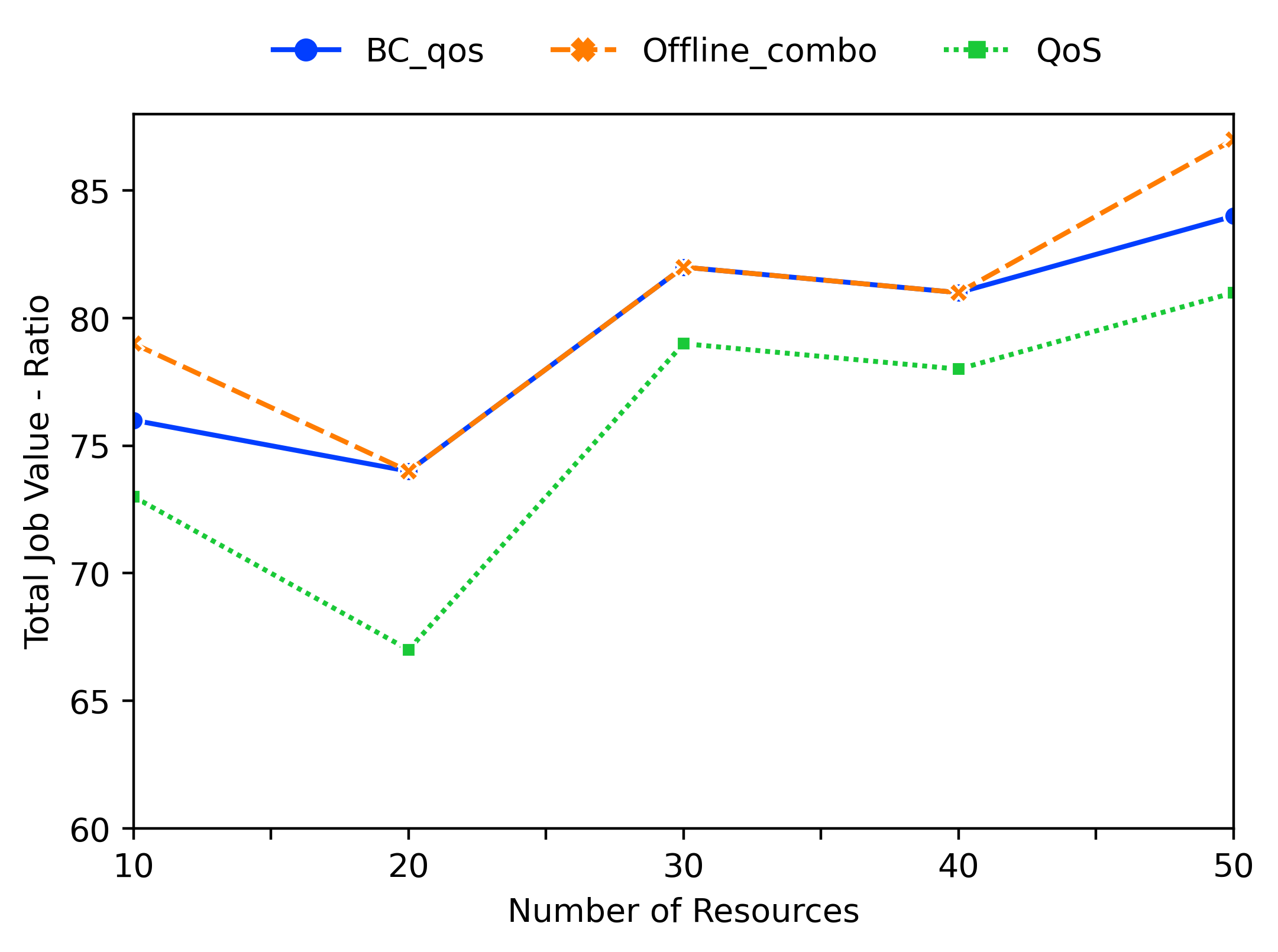}
\caption{Comparing Offline Learning and Behaviour Cloning with combination of heuristic policies.} 
\label{fig:Offline_BC_best}
\end{figure}

\textit{Analysis}: Figure \ref{fig:Offline_RL_10Res} shows the performance of Offline RL compared to BC, with the replay buffer of prior data consisting of a mix of well-performing and poor transitions. The Offline\_combo performs significantly better than BC\_combo when the training data has a mix of good and bad transitions. From the figure \ref{fig:Offline_RL_10Res}, we note that BC\_qos agent performs better than BC\_qos\_sjf and BC\_combo. The BC\_qos\_sjf and BC\_combo agents were trained with a mix of poor-quality transitions. Since the BC agent learns to merely mimic the transitions in the replay buffer, the presence of poor-quality transitions affects the performance of these two agents. 

Figure \ref{fig:Offline_RL_scale} shows the performance of Offline RL compared to BC and Qos heuristic methods when scaled to 50 resources. At scale, the Offline\_combo significantly outperforms both BC\_combo and QoS policies. Additionally, Figure \ref{fig:Offline_BC_best} shows that Offline\_combo agent trained on sufficiently noisy suboptimal data can attain at-par or better performance than even BC\_qos method trained with expert demos. Additionally, the Offline\_combo, Offline\_qos\_sjf, and Offline\_qos agents perform similarly (figure \ref{fig:Offline_RL_10Res}). The Offline RL methods benefit from stitching parts of suboptimal trajectories. For example, suppose the dataset contains a subsequence illustrating arrival at state $x+1$ from state $x$ and another subsequence illustrating arrival at state $x+2$ from state $x+1$. In that case, an effective Offline RL method should be able to learn how to arrive at state $x+2$ from state $x$, which might provide a substantially higher final reward than any of the subsequences in the dataset. This sort of ``transitive induction" occurs on a portion of the state variables, effectively inferring potentially optimal behavior from highly suboptimal components.

The Offline RL method facilitates generalization, i.e., it can be adapted to learn from any policy (e.g., Oracle, heuristic policies) that optimizes a specific objective, such as job value, resource utilization, or energy efficiency. We remark that the Offline RL method is significantly better when training buffers are loaded with a mix of transitions generated by more than one policy. With mixed transitions (well-performing and poor), Offline RL can learn a general policy, possibly stitching multiple policies to get a better one. 

\subsection{Online Learning}
The most obvious challenge with Offline RL is that because the learning algorithm must entirely rely on the static dataset, there is no possibility of improving by exploration. If the dataset does not include transitions that illustrate high-reward regions of the state space, it may be impossible to learn such high-reward regions. For example, when the dataset size is limited, some learning algorithms tend to overfit on the small dataset, or if the dataset state-action distribution is biased, neural network training may only provide brittle, non-generalizable solutions. Additionally, not all environments have historical or high-quality datasets that can be readily used for training. 

In Online learning, the agent interacts with the environment and explores numerous state-action pairs to learn a generalizable policy. Model-free deep RL methods are notoriously expensive in terms of their sample complexity. Even relatively simple tasks can require millions of data collection steps, and complex behaviors with high-dimensional observations might need substantially more. What is considered an upside, the exploration process, is also the downside because the exploration process is time-consuming, where the agent alternates between the exploration and exploitation phases to learn decent policy. We trained the BC and Offline agents for 500k steps and the Online agent for 1 million steps. Everything else being the same (e.g., neural net configuration, batch size), the Online agent was trained twice as long as the BC and Offline methods to achieve similar performance goals. 

\begin{figure}[tbh]
\centering
\includegraphics[scale=0.50]{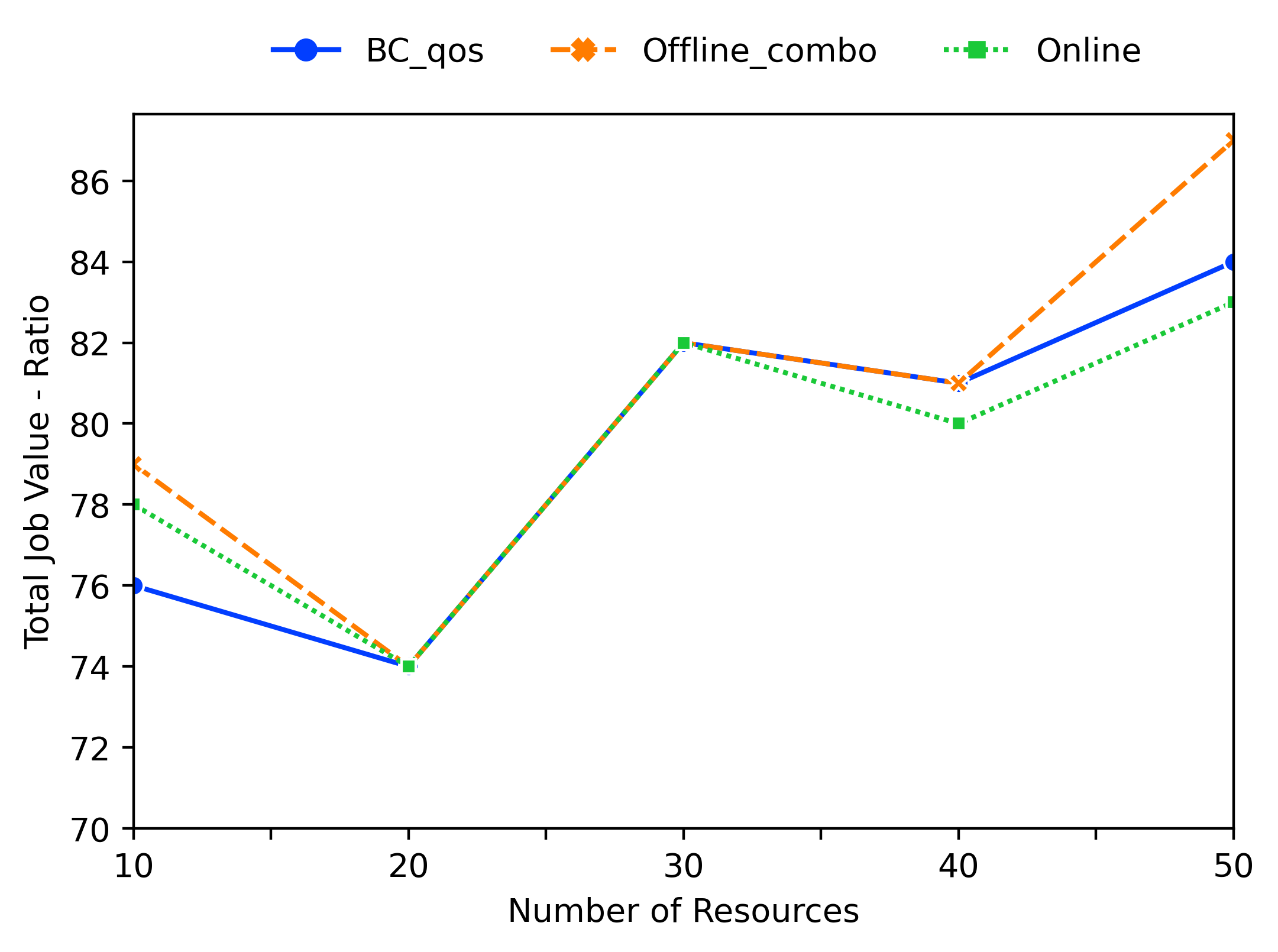}
\caption{Comparing the performance of BC, Offline and Online Learning methods.} 
\label{fig:BC_offline_online_comp}
\end{figure}

\textit{Analysis}: Figure \ref{fig:BC_offline_online_comp} shows the performance of Online RL compared to BC\_qos and Offline\_combo Learning methods. We note that the Online method performs as well as the BC and Offline methods when the state space is small (10-30 resources). The performance drops as the state-space increases (40-50 resources). This is because exploring bigger state space needs more interactions with the environment and, therefore, a longer time to train. With enough time and random starting points, the online agent might eventually match or outperform the BC and Offline methods for large state-space problems. We will explore this future in our future work. 

\subsection{Combining Offline and Online Learning}
Online RL provides an appealing formalism for learning control policies from experience. However, the classic active interactions with the environment require a lengthy active exploration process for each behavior. Suppose we allow RL algorithms to use historical datasets to aid online learning effectively. In that case, the learning process could be made substantially more practical: the prior data would provide a starting point, a \textbf{launchpad}, that mitigates challenges due to exploration and sample complexity, while the online training enables the agent to perfect the desired skill. Such prior data could either constitute expert demonstrations or, more generally, sub-optimal prior data that illustrates potentially useful transitions. Given the dataset, $\mathcal{D}$, of transitions generated by a heuristic policy, our goal is to leverage $\mathcal{D}$ for pre-training and use some number of online interactions to learn the good generalizable scheduling policy.

For this experiment, we used a modified version of advantage weighted actor-critic (AWAC)~\cite{nair2021awac}(Algorithm \ref{algo:online_training}) framework, which enables rapid learning of skills with a combination of prior datasets and online experience. This framework leverages offline data and quickly performs online fine-tuning of RL policies. Additionally, incorporating prior datasets can reduce training time while exploring large state-space.

\begin{figure}[tbh]
\centering
\includegraphics[scale=0.50]{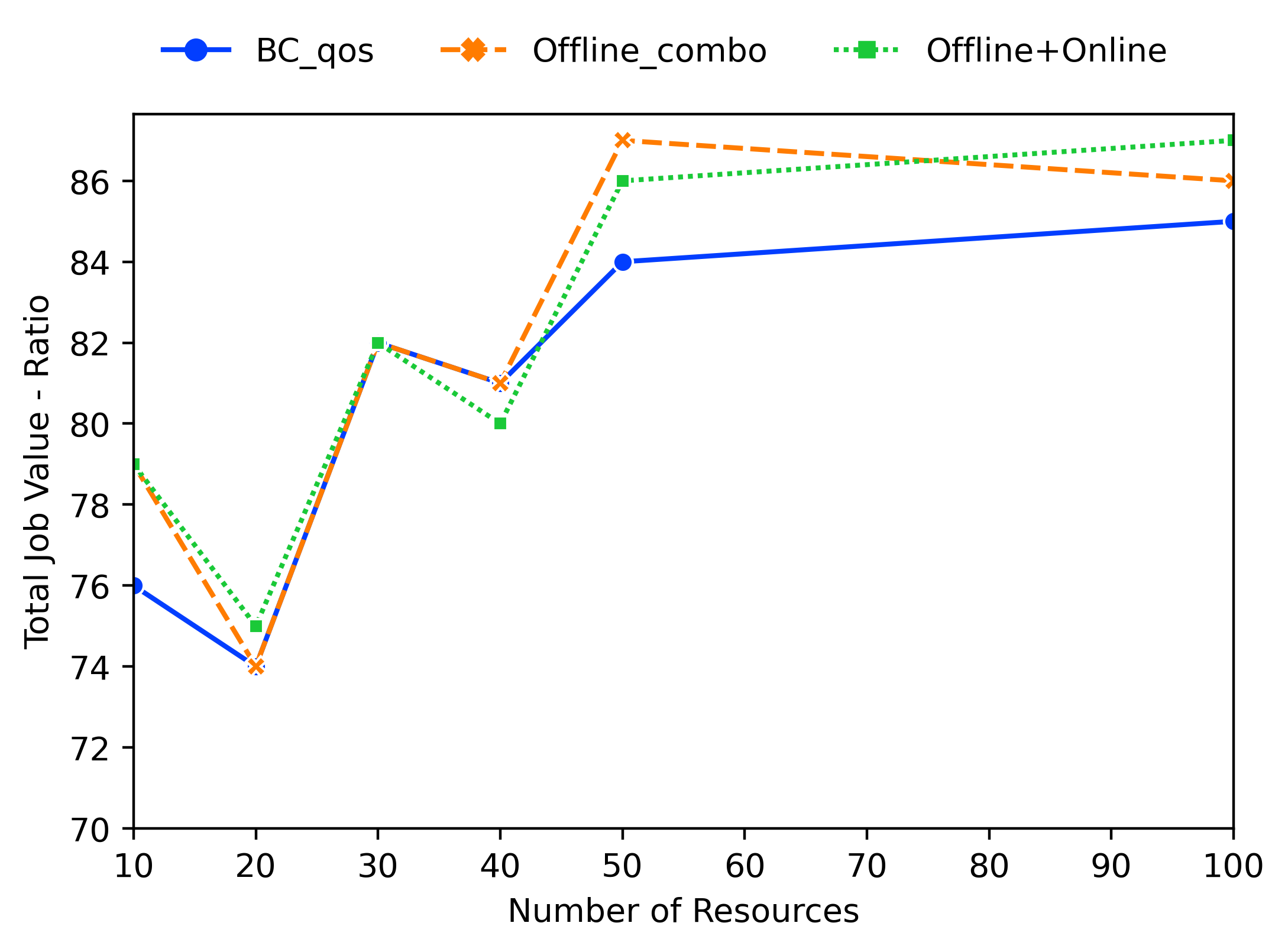}
\caption{Comparing BC, Offline Learning and combining Online Learning with Offline data.} 
\label{fig:BC_Offline_and_Online_Comp}
\end{figure}

\textit{Analysis}: Figure \ref{fig:BC_Offline_and_Online_Comp} shows the performance of Online training using Offline data (Offline+Online) compared to BC\_qos and Offline\_combo learning methods. For this experiment, we trained the agent for 50k Offline steps (pre-loading QoS transitions to the replay buffer) and 1 million online steps. Even though the Offline+Online was trained for merely 50k offline steps, the Offline+Online agent's performance is at par or better than either BC or Offline methods for large problem sizes (100 resources). This framework is effective for pre-training from off-policy datasets but is also well suited to continuous improvement with online data collection. Additionally, this framework can utilize different types of prior data: demonstrations, suboptimal data, or random exploration data.

\subsection{Effectiveness of Offline and Online methods}
Next, we will discuss the effectiveness of the Offline and Online learning methods and present the scenarios under which a given RL technique is applicable.

First, we collected 80 rollouts (100-200k samples) of offline experience data where the heuristic scheduling policies (QoS, SJF\_QoS, and Combo) select the actions. Second, we load the offline experience into the empty replay buffer during the RL scheduler's training. We trained three RL scheduler agents with BC, Offline RL, and Offline+Online methods. In each of these methods, the RL scheduler's actor net learns to mimic the action choices of the heuristic data in its replay buffer to varying degrees. To evaluate the success of the learning process, we simulated new rollouts controlled by the original heuristic and measured the percentage of steps where the policy's action is equal to the heuristic's decision in the current state. The \textit{action agreement} metric provides insight into RL method's ability to learn heuristic policies and its affect the performance. 

\begin{table}[tbh]
\begin{center}
    \begin{tabular}{P{3.3cm} | P{1.0cm}  P{1.0cm}  P{1.3cm}  P{1.0cm}}
        \hline
        Action Agreement & QoS & SJF\_QoS & Combo \\
        \hline
        BC & $95\%$ & $99\%$ & $98\%$ \\
        \hline
        Offline & $91\%$ & $91\%$ & $87\%$ \\
        \hline
        Offline+Online & $21\%$ & $21\%$ & $17\%$ \\
        \hline
    \end{tabular}
    \newline \newline
    \caption{Action agreement of BC, Offline and Offline+Online learning methods}
    \label{table:behavioral_cloning}
\end{center}
\end{table}

\textit{Analysis}: The action agreement percentage for $10$ resource environment are shown in Table \ref{table:behavioral_cloning}. The BC policy, trained with SJF\_QoS and Combo datasets, has the highest action agreement ($99\%$ and $98\%$ respectively), confirming that the BC learning method directly mimics the underlying policy in the dataset. When the replay buffer dataset consists of multiple policies (BC\_qos\_sjf and BC\_combo), it is more likely that BC will mimic the actions of one of the underlying policies attributing to the high action agreement. When trained using a dataset with only QoS policy, BC\_qos scheduler agrees with the QoS policy $95\%$ of the time. The remaining $5\%$ of the time BC is likely generating random actions. On the other hand, the offline RL scheduler demonstrates a lower action agreement with the policy in the dataset. When trained using a dataset with QoS and SJF\_QoS policies, the Offline RL agent agrees $91\%$ of the time. The offline RL agent's action agreement drops to $87\%$ with combo datasets confirming that offline RL is resilient to low-quality datasets and extracts good policies even when the noise-to-signal ratio is high. The Offline+Online method, pre-trained with offline data and using online interactions for improvement, shows $21\%$ action agreement when pre-trained with QoS and SJF\_QoS and $17\%$ with combo data. This low action agreement indicates that the Offline+Online method learns policies significantly different from the underlying heuristic policies leading to better performance.

\begin{table}[tbh]
    \begin{tabular}{P{1.8cm} | P{0.5cm} P{0.7cm} P{2.0cm} P{0.7cm}}
    \hline
        Data Quality & BC &  Offline &  Offline+Online & Online \\
        \hline
        High  & \checkmark & \checkmark & \checkmark & NA \\
        \hline
        Medium  & $\times$ & \checkmark & \checkmark & NA \\
        \hline
        Low  & $\times$ & $\times$ & \checkmark & NA \\
        \hline
        No data & $\times$ & $\times$ & $\times$ & \checkmark \\
    \hline
\end{tabular}
\newline \newline
\caption{Choosing between BC, Offline and Offline+Online learning methods}
\label{table:recommendation}
\end{table}

Finally, Table \ref{table:recommendation} presents our recommendations on which RL method is suitable based on the quality of prior datasets and learning environments. Suppose the practitioners have access to high-quality datasets (expert demonstrations) and want to extract the general underlying policy. In that case, BC is a better choice, although Offline methods will also perform equally well. One caveat is that if the expert demos are generated by a highly customized policy for a specific domain, then BC might fail to generalize well in a modified domain. If the prior datasets are of mixed quality (expert demonstrations mixed with suboptimal transitions), then Offline and Offline+Online RL methods are suitable. Offline RL is more resilient to noise but needs some good transitions closer to high-reward areas. 

Similarly, the Offline+Online method is resilient to noisy datasets, but this method also has the advantage of interacting with the environment to improve the pre-trained policy. Therefore, the Offline+Online method can be used even when the noise-to-signal ratio in the prior dataset is high. The agent (pre-trained with a noisy dataset) will eventually learn good policies after some interactions with the environment. Finally, if practitioners have access to prior datasets or the learning environment is novel (e.g., to learn new actions to avoid QoS violations in green datacenters), then the Online RL method is most suitable. In terms of training time, BC training is generally faster because the supervised actor update is easier to learn than the Q function (Offline learning). For the same reason, Offline learning is faster than Offline+Online learning. We observed that, for our problem setting, Online learning took the longest to learn reasonable policies, given all other parameters being the same. 

\section {Future work} \label{future_work}

In future work, we would like to empirically analyze the performance of offline learning with diverse data coming from different workloads. The goal of data-driven methods is to train a model that attains good performance on data coming from the same distribution as the training data. In offline RL, the basic idea is to learn a policy that does something differently (likely better) from the behavior pattern observed in dataset $\mathcal{D}$. Distributional shift issues can be addressed in several ways, and we would like to explore this further. Additionally, we would like to analyze the effect of the size of the prior dataset for offline learning, training duration, and performance with various workloads.  

\section{Related Work} \label{related_work}


\textbf{Online RL Schedulers}. The authors in ~\cite{DBLP:Decima} used graph convolutional embedding for cluster job scheduling in an online setting where the jobs had a DAG structure that could be exploited. The scheduler in~\cite{RLScheduler-SC20} implements a co-scheduling algorithm using online RL by combining application profiling and cluster monitoring. In ~\cite{Vana_cocorl_2023}, the authors propose a job scheduling scheme based on constraint-controlled online RL implementation where the scheduler agent's learning objective is to maximize the total revenue while minimizing job delays in green datacenters. Another work~\cite{datacenter-energy-cdc} presents a unified management approach for the thermal and workload distribution in datacenters, modeled as a Model Predictive Control problem. Much of the existing work applies online RL methods, completely ignoring the benefits of data-driven learning methods. 

\textbf{Behavior Cloning and Offline RL}. The authors in ~\cite{anish_il_2020} propose scheduling on Domain-specific systems-on-chip as a classification problem and propose a hierarchical imitation learning based scheduler that learns from an Oracle to maximize the performance of multiple domain-specific applications. A similar framework is suggested in ~\cite{mandal-2019} for mobile platforms. HiLITE~\cite{Hilite-2020} employs a hierarchical imitation learning framework to maximize energy efficiency while satisfying soft real-time constraints on embedded Systems on Chip. To the best of our knowledge, none of the existing research applies offline RL methods or empirically demonstrate when to use the various offline RL methods for job scheduling problem. In this paper, we aimed to understand if, when, and why offline RL is a better approach for tackling sequential decision-making problems, specifically job scheduling in green datacenters.

\section{Conclusion} \label{conclusion}


We explored two data-driven RL methods, namely 1) Behaviour Cloning and 2) Offline RL, which aim to learn policies from logged data without further interaction with the environment. These methods address the challenges concerning the data collection costs and safety, particularly pertinent to real-world applications of RL. Although offline learning methods produce good results, we showed that the performance is highly dependent on the quality of the historical datasets. Finally, we utilize offline RL as a \textbf{launchpad} to learn effective scheduling policies from a prior dataset collected using expert demonstrations or heuristic policies. By effectively incorporating prior datasets to pre-train the RL agent, we short-circuit the random exploration phase to learn reasonable policies with online learning. The contribution of this study is not just another RL scheduler but a systematic study of what makes sense - standard offline RL, pure Online RL methods, or offline pre-training with online fine-tuning. 

\bibliography{icaps-latex-2023}


\end{document}